\documentclass[10pt,twocolumn,letterpaper]{article}

\usepackage{cvpr}              
\usepackage{multirow}
\usepackage[dvipsnames]{xcolor}
\usepackage[normalem]{ulem}

\usepackage{algorithm}
\usepackage{algpseudocode}

\usepackage{amsmath}
\usepackage{amssymb}
\usepackage{amsfonts}

\definecolor{cvprblue}{rgb}{0.21,0.49,0.74}
\usepackage[pagebackref,breaklinks,colorlinks,allcolors=cvprblue]{hyperref}

\def\paperID{} 
\def\confName{CVPR}
\def\confYear{2026}

\title{Physically Realistic Sequence-Level Adversarial Clothing\\for Robust Human-Detection Evasion}

\author{
Dingkun Zhou\textsuperscript{1} 
Patrick P. K. Chan\textsuperscript{1,\dag} 
Hengxu Wu\textsuperscript{2} 
Shikang Zheng\textsuperscript{1} 
Ruiqi Huang\textsuperscript{1} 
Yuanjie Zhao\textsuperscript{1}\\
\textsuperscript{1}School of Future Technology, South China University of Technology, Guangzhou, China\\
\textsuperscript{2}Institute for Interdisciplinary Information Sciences, Tsinghua University, Beijing, China\\
{\small\texttt{jackkun818@gmail.com}} \quad
{\small\texttt{patrickchan@scut.edu.cn}} \\
{\small\texttt{wuhx24@mails.tsinghua.edu.cn, zhengshikang@berkeley.edu}} \\
{\small\texttt{\{202364870722, 202364871302\}@mail.scut.edu.cn}}
}
\begin{document}



\maketitle

\begin{abstract}
Deep neural networks used for human detection are highly vulnerable to adversarial manipulation, creating safety and privacy risks in real surveillance environments. Wearable attacks offer a realistic threat model, yet existing approaches usually optimize textures frame by frame and therefore fail to maintain concealment across long video sequences with motion, pose changes, and garment deformation. In this work, a sequence-level optimization framework is introduced to generate natural, printable adversarial textures for shirts, trousers, and hats that remain effective throughout entire walking videos in both digital and physical settings. Product images are first mapped to UV space and converted into a compact palette and control-point parameterization, with ICC locking to keep all colors printable. A physically based human–garment pipeline is then employed to simulate motion, multi-angle camera viewpoints, cloth dynamics, and illumination variation. An expectation-over-transformation objective with temporal weighting is used to optimize the control points so that detection confidence is minimized across whole sequences. Extensive experiments demonstrate strong and stable concealment, high robustness to viewpoint changes, and superior cross-model transferability. Physical garments produced with sublimation printing achieve reliable suppression under indoor and outdoor recordings, confirming real-world feasibility.
\end{abstract}    
\section{Introduction}

Deep neural networks (DNNs) have achieved strong performance in human detection and are now widely used in surveillance~\cite{Shah_2021_ICCVW}, autonomous driving~\cite{Jiang_2024_AppliedSciences}, and person re-identification~\cite{Schreier_2023_ICCVW}. Despite this progress, numerous studies~\cite{Yu2024PhysicalAdversarialSurvey,Costa2023ADLSurvey, chan2024evasion} have shown that DNNs remain highly vulnerable to adversarial perturbations. Even small pixel-level changes or subtle texture manipulations can trigger severe detection failures~\cite{wang2021das}, which creates concrete safety and privacy risks once models are deployed in real environments.

\begin{figure}[t]
  \centering
  \begin{subfigure}{0.589\linewidth}
    \centering
    \includegraphics[width=\linewidth]{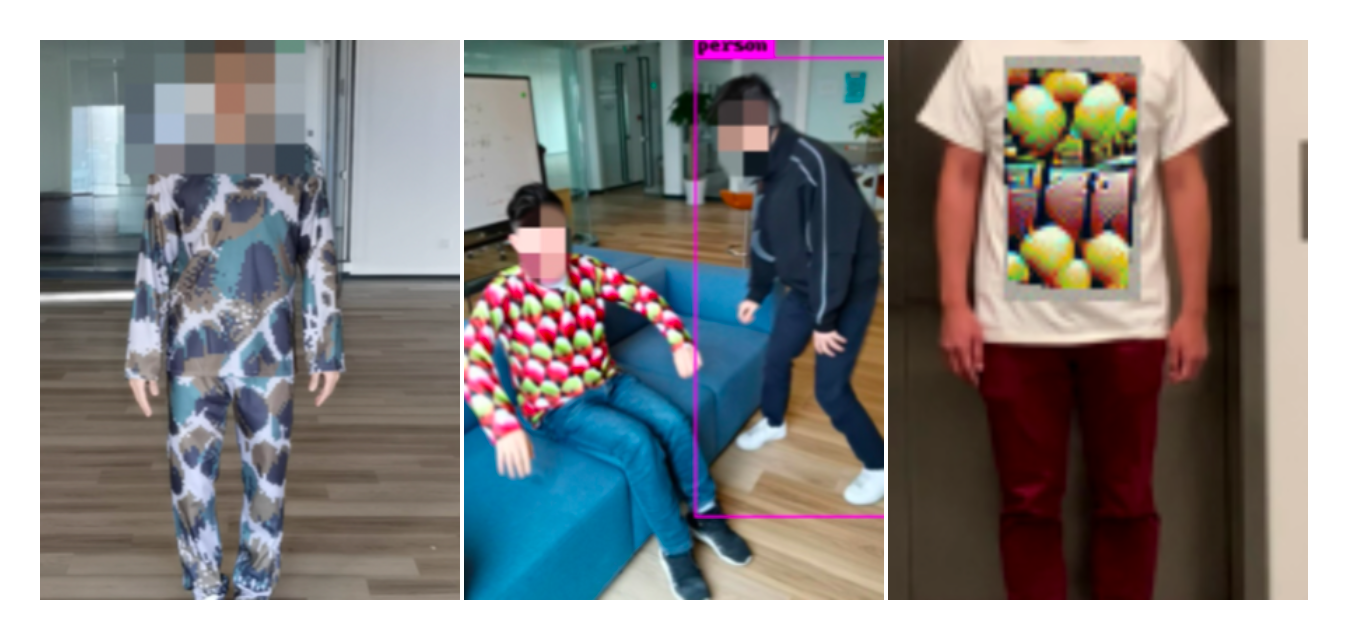}
    \subcaption{State-of-the-art Wearable Attacks}
  \end{subfigure}\hfill
  \begin{subfigure}{0.411\linewidth}
    \centering
    \includegraphics[width=\linewidth]{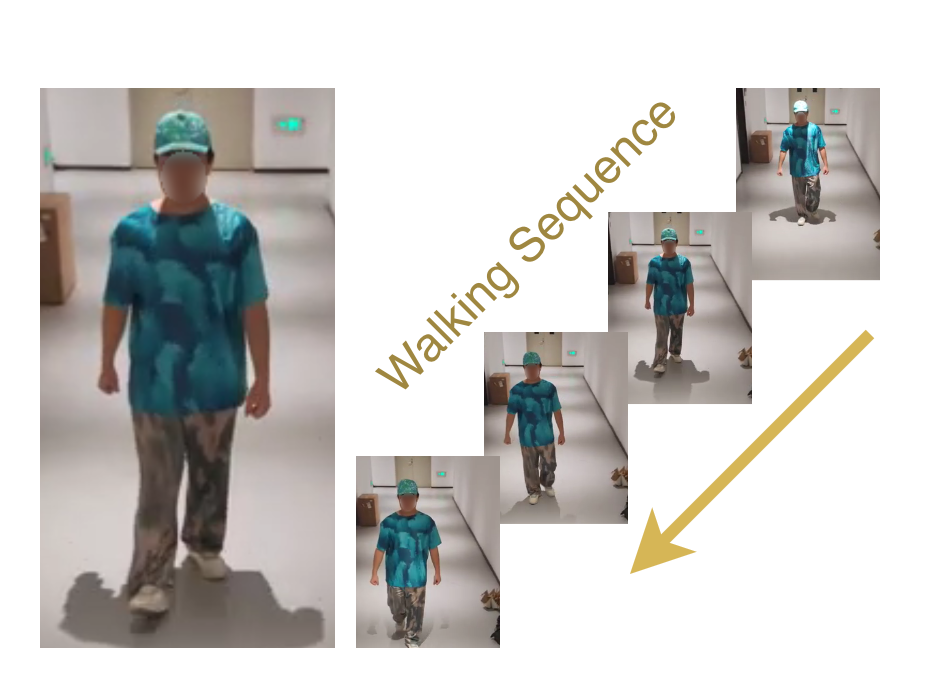}
    \subcaption{ours}
  \end{subfigure}
  \vspace{-5mm}
  \caption{Existing wearable attacks versus our approach. Prior attacks generally target static frames, often rely on visually conspicuous patterns, and do not maintain concealment throughout motion. In contrast, our method produces natural full-outfit adversarial clothing and achieves stable evasion across entire walking sequences, enabling reliable video-level attacks.}
  \label{fig:intro}
\end{figure}

Adversarial attacks realized through clothing have received increasing attention because garments are natural, wearable, and visually unobtrusive~\cite{hu2022advtexture,hu2023advcamocloth,chan2023learning}. However, existing adversarial clothing methods are usually optimized on individual frames and only over a narrow range of viewpoints. As a result, their performance degrades rapidly under continuous human motion, elevation changes, garment deformation, and shifting camera geometry~\cite{xu2020advTshirt,advPattern2019}. These limitations highlight that reliable real-world evasion requires robustness across entire video sequences rather than isolated frames, and that clothing textures must be jointly optimized within a unified framework that accounts for scene dynamics.

In this work, a sequence-level adversarial clothing framework is proposed to generate natural textures for shirts, trousers, and hats that can mislead human detectors in both digital and physical settings. The attack is formulated as an optimization problem where printable UV textures are learned to reduce the detector's confidence and intersection-over-union (IoU)~\cite{jaccard1912} across a complete walking sequence under varying viewpoints, poses, illumination, material properties, and backgrounds. A unified pipeline (Fig.~\ref{fig:pipeline}) is introduced. Product images are first projected into UV space to obtain texture initialization and a dual-domain K-Means~\cite{lloyd1982least} parameterization that yields a printer-safe color palette and a compact set of spatial control points. The control points are then reconstructed into textures through a differentiable generator. Human motion is synthesized by interpolating key poses and realistic garment deformation is produced using a pretrained neural cloth simulator with randomized material parameters. The textured garments are rendered under diverse camera and lighting settings, and a sequence-level loss is computed using temporally weighted detection confidences. Control points are finally refined by backpropagating through the generator, renderer, and simulator so that the optimized textures remain effective throughout the entire sequence.

The proposed approach produces adversarial garments that demonstrate significantly stronger robustness than existing methods. Digital experiments show high SeqASR, low CVaR, and high NDR across a wide range of camera elevations, and strong cross-model transferability is achieved on five major detectors. Physical evaluations with fabricated garments further confirm that the optimized textures remain effective under real-world motion, deformation, and illumination. These results indicate that practical sequence-level adversarial robustness can be achieved when clothing textures are optimized jointly with physical dynamics and full-sequence temporal constraints.

In summary, our main contributions are as follows:
\begin{itemize}
    \setlength{\itemsep}{0.5em} 
    \item A sequence-level adversarial clothing framework is introduced, in which textures for upper garments, trousers, and hats are optimized jointly to suppress human-detector confidence across entire video sequences rather than individual frames. The framework incorporates physical garment simulation, camera–motion diversity, and printer-safe texture parameterization, enabling textures that remain effective under realistic dynamics.
    \item A physically grounded deformable-garment pipeline is developed that integrates HOOD-based cloth simulation, UV-domain color-palette locking, and a dual-domain control-point representation. This design ensures that adversarial textures remain printable, physically plausible, and robust to pose-dependent deformations, illumination changes, and motion.
    \item A video-level evaluation protocol is established using the proposed SeqASR, CVaR, and NDR metrics to quantify temporal stability and worst-case exposure. Extensive digital and physical experiments demonstrate that the proposed approach achieves higher sequence robustness, stronger cross-model transferability, and greater physical reliability than existing state-of-the-art methods.
\end{itemize}

\section{Related Work}
\label{sec:related-work}

Early work showed that small adversarial perturbations can fool deep neural networks. Physical attacks later extended these ideas to object detection, with patch-based and texture-based methods forming two major categories.

\noindent\textbf{Patch-based Physical Attacks} place optimized regions onto an object to suppress or redirect detector outputs~\cite{brown2017adversarialpatch,eykholt2018rp2,thys2019fooling,AdvGAN,he2024dorpatch,guesmi2023dap,xu2020advTshirt}. Classic universal patches~\cite{brown2017adversarialpatch} use EOT~\cite{EOT} for robustness, and pedestrian-evasion patches have been shown under surveillance settings~\cite{thys2019fooling}. Later works improved realism using generative priors~\cite{AdvGAN}, robustness through sparse or distributed placement~\cite{he2024dorpatch}, or deformation-aware modeling for non-rigid surfaces~\cite{guesmi2023dap}.
Despite progress, patch attacks remain limited in viewpoint robustness, often appear visually conspicuous, and typically degrade under clothing deformation or long-term motion.

\noindent\textbf{Texture-based Adversarial Garments} methods optimize full-surface UV textures for stronger multi-view robustness~\cite{pestana2022transferable3dtex,yao2020multiview3dattack,wang2021das,hu2022advtexture,hu2023advcamocloth,lu2024advocl}. Differentiable rendering enables end-to-end viewpoint-aware optimization~\cite{pestana2022transferable3dtex}, and multi-view constraints improve performance under camera-pose changes~\cite{yao2020multiview3dattack}. For adversarial clothing, Hu et al.~\cite{hu2022advtexture,hu2023advcamocloth} generated physically printable textures with natural camouflage designs and validated robustness under real garments and multi-angle views. Diffusion-based methods~\cite{lu2024advocl} further enhanced realism under occlusion.
However, existing texture attacks still optimize at the frame level, making them sensitive to garment dynamics, long sequences, and material variation, and they exhibit limited cross-model transferability.

\noindent\textbf{Positioning of Our Work:} Our method addresses these gaps through sequence-level optimization, physically grounded garment dynamics, and a compact, printer-safe control-point parameterization. This enables stable evasion across long motion sequences, materials, and viewpoints in both digital and physical settings.

\section{Sequence-Level Adversarial Clothing}

\begin{figure*}[t]
  \centering
  \includegraphics[width=\linewidth]{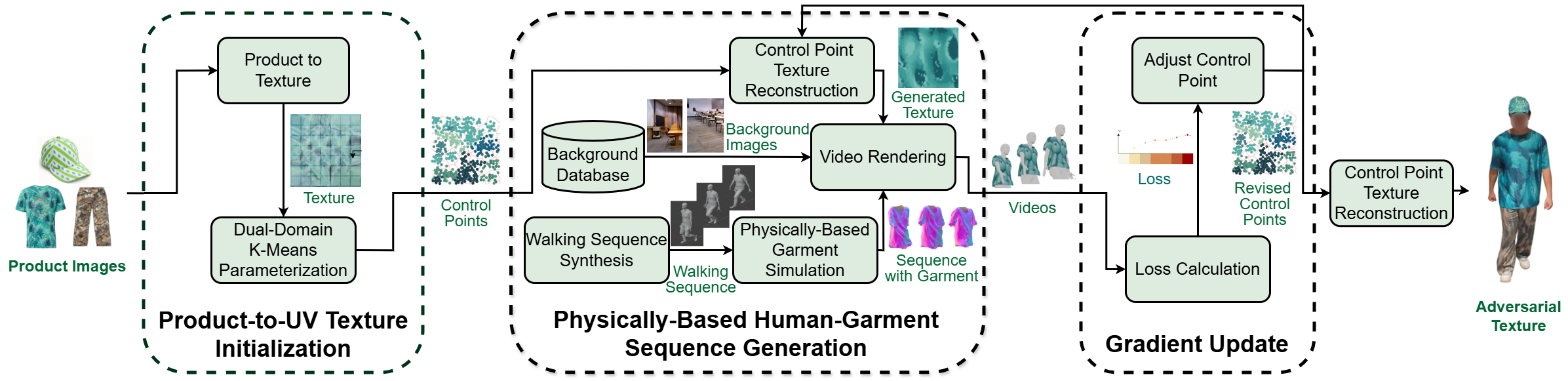}
  \caption{Main pipeline of our proposed model.}
  \label{fig:pipeline}
\end{figure*}

The overall pipeline (Fig.~\ref{fig:pipeline}) converts natural product images into physically valid adversarial garment textures through a sequence of differentiable stages. In the \emph{product-to-UV initialization} step, each product image is mapped to the canonical UV domain via Pix2Surf~\cite{Mir_2020_CVPR}, then compressed using a dual-domain K-Means~\cite{lloyd1982least} that extracts a printer-safe color palette and spatial control points forming a low-dimensional texture representation.

In \emph{physically-based human–garment sequence generation}, the control-point texture is reconstructed and applied to a synthesized walking sequence. Realistic cloth dynamics are simulated using a pretrained HOOD-based physical propagator~\cite{Grigorev_2023_CVPR} under randomized fabric parameters, after which the textured garment is rendered under diverse camera, lighting, and background conditions.

Finally, \emph{sequence-level control-point optimization} refines the control points using expectation-over-transformation (EOT)~\cite{EOT}. A sequence-level loss enforces temporal robustness across entire motion cycles, while a repulsive regularizer prevents control-point clustering in UV space. Through iterative rendering and gradient updates, the pipeline yields a physically plausible, printable adversarial texture that remains effective across varied motions and viewpoints.

\subsection{Product-to-UV Texture Initialization}
\label{sec:method:init}

\subsubsection{Product to Texture} 
Given natural product images, we first map them into the canonical UV domain. For each garment category $g \in \{\mathrm{upper}, \mathrm{lower}, \mathrm{hat}\}$ with its corresponding product image $I_g$, a frozen Pix2Surf~\cite{Mir_2020_CVPR} encoder outputs a dense correspondence from image pixels in $I_g$ to UV coordinates on $\mathcal{M}_g$ and a per-pixel visibility mask indicating garment regions, where $\phi_g: \mathcal{M}_g \rightarrow [0,1]^2$ defines the mapping from the 3D template surface $\mathcal{M}_g$ to the 2D \emph{UV} texture domain, and U and V denote the horizontal and vertical axes of the texture plane, respectively. Using this correspondence, standard UV baking is performed on a UV grid of size $H \times W$. For pixels covered by valid correspondences, color values are bilinearly sampled from $I_g$ and written into the initial UV texture $T_g^0 \in \mathbb{R}^{H \times W \times 3}$. In parallel, a UV validity mask $V_g \in \{0,1\}^{H \times W}$ is constructed, where $V_g(u,v)=1$ indicates valid UV islands. 

\subsubsection{Dual-Domain K-Means Parameterization}

This module transforms the initial texture $T_g^0$ into a low-dimensional, printer-safe, and differentiable representation through two sequential K-Means~\cite{lloyd1982least} stages: palette extraction and spatial control-point extraction~\cite{zhou2025balancing}.

\noindent\textbf{Stage 1: K-Means for Palette Extraction.}
Pixels of $T_g^0$ are clustered in the sRGB 8-bit color space using K-Means~\cite{lloyd1982least} to form a compact and semantically meaningful color palette. Each pixel $c_g(x,y)\!\in\![0,255]^3$ is assigned to one of $K$ clusters indexed by $c$, with $\mu_{g,c}$ denoting the centroid color. A larger $K$ increases color richness and captures subtle material details, while smaller $K$ values limit expressiveness; however, computation scales linearly with $K$, and overly large values may cause overfitting and noise. To ensure printability, each centroid $\mu_{g,c}$ is passed through an ICC-based RGB$\leftrightarrow$CMYK round-trip conversion, $\hat{\mu}_{g,c}=\Gamma_{\text{rgb}\leftarrow\text{cmyk}}\!\big(\Gamma_{\text{cmyk}\leftarrow\text{rgb}}(\mu_{g,c})\big)$, where $\Gamma$ denotes calibrated color-space mappings from ICC profiles. The resulting locked palette $\hat{K}_g=\{\hat{\mu}_{g,c}\}_{c=1}^{K}$ constrains all colors within the printer gamut, and each texture pixel is represented as a convex combination of palette colors, ensuring gamut consistency without extra penalties.

\noindent\textbf{Stage 2: K-Means for Control-Point Extraction.}
To compactly encode spatial structure, K-Means is applied to the UV coordinates $X^{g}_{c}$ of pixels grouped by the locked color palette $\hat{K}_g$, where $c$ indexes the color clusters $\hat{\mu}_{g,c} \in \hat{K}_g$. This process generates spatial control points $p_{g,c,j}$, where $j$ denotes the index of the control point within the $c$-th color cluster. 
All control points of garment $g$ form the control-point set 
$P_g = \{p_{g,c,j}\mid c=1,\ldots,K,\ j=1,\ldots,P_{\max}\}$. 
A uniform upper bound $P_{\max}$ is imposed on the number of control points per cluster, and interpolation is applied when fewer points are obtained to maintain consistent spatial density across the UV domain.


\subsection{Physically Based Human-Garment Generation}

\subsubsection{Control-Point Texture Reconstruction}
The garment texture $T_g^P$ is reconstructed from the control-point set $P_g$ using the differentiable texture generation process ~\cite{hu2023advcamocloth}, i.e., $P_g \mapsto T_g^P$. Control points are first mapped to a channel logit field, which is perturbed by Gumbel noise to preserve stochasticity and avoid gradient saturation. The noisy logits are then transformed by a Gumbel–Softmax~\cite{Jang2017GumbelSoftmax,maddison2017the} function into normalized mixture coefficients, which are convexly combined with the ICC-projected palette to produce a printable and differentiable optimizable texture.

\subsubsection{Walking Sequence Synthesis}

The human walking sequence is synthesized using Blender software~\cite{blender42lts}. Assuming a number of key poses of a standard-size male, each containing full-body joint rotations and global translations in canonical space, intermediate frames are generated by interpolating SMPL pose data~\cite{SMPL:2015} between these key poses. This interpolation produces a smooth and temporally coherent walking sequence $\mathcal{P} = \{p_t\}_{t=1}^{T}$, which serves as the input for subsequent garment simulation. The details are given in the Appendix.


\subsubsection{Physically-Based Garment Simulation}


Garment dynamics for a short-sleeved upper garment and long trousers are simulated under physical constraints to generate realistic deformations corresponding to the walking sequence $\mathcal{P}$, while the hat is modeled as a rigid dome without physical deformation. A pretrained physical propagator $f_\psi$, based on the open-source HOOD framework~\cite{Grigorev_2023_CVPR}, is employed to advance the garment geometry $X_\tau^{\,g}$ over time, producing the deformation sequence $\mathcal{X}_g=\{X_\tau^{\,g}\}_{\tau=0}^{t}$. Several combinations of material parameters $\Phi_g$, including fabric softness, bending stiffness, and density, are considered to control the dynamic response of different garment types. The deformation sequence represents frame-wise garment geometries that capture realistic cloth bending, stretching, and contact interactions throughout the motion, computed through a stable integration process under physical constraints. To further enhance physical plausibility, we introduce a fixed-point anchoring strategy for high-friction or stitched regions such as necklines, shoulder seams, and waistbands, where selected vertices are rigidly attached to corresponding body points. This mechanism effectively prevents drift and sliding, improves numerical stability, and ensures consistent deformation behavior across long motion sequences.

\subsubsection{Video Rendering}


The projected texture $T_g^P$ is rendered on the physically simulated garment deformation sequence $\mathcal{X}_g$ using a differentiable renderer~\cite{ravi2020pytorch3d}. Each rendering scene is parameterized by a set of controllable variables $\{\mathcal{C}, \mathcal{O}, \mathcal{I}, \mathcal{B}\}$ that define the camera configuration, human motion, illumination, and background. The camera parameters $\mathcal{C}$ include view angle, distance, and elevation, enabling the synthesis of multiple viewpoints. The motion parameters $\mathcal{O}$ describe the walking direction, speed, and starting position of the subject, introducing temporal and spatial variation across the sequence. The imaging parameters $\mathcal{I}$ specify lighting intensity, color temperature, and direction, allowing simulation of diverse illumination settings. A static outdoor background, parameterized by $\mathcal{B}$, is used by default to ensure consistent brightness, while alternative backgrounds can be substituted to evaluate the robustness of the adversarial texture under different visual contexts.

%
%
%
%

%
%

\subsection{Sequence-Level Control-Point Optimization}

The optimal control-point set $P_g^*$ is obtained in the framework of the expectation-over-transformation (EOT) optimization:
\begin{equation}
\min_{P_g}\ \mathbb{E}_{\mathcal{C}, \mathcal{O}, \mathcal{I}, \mathcal{B}, \Phi_g}\!\left[ L(x) \right],
\end{equation}
where $x$ denotes a rendered frame generated under the scene parameters $\{\mathcal{C}, \mathcal{O}, \mathcal{I}, \mathcal{B}, \Phi_g\}$. 
Following previous work~\cite{hu2023advcamocloth}, the loss function $L$ represents the detector’s objectness confidence corresponding to the predicted person bounding box that achieves the highest intersection-over-union (IoU) ~\cite{jaccard1912} with the ground-truth box, producing a scalar loss for each frame. 

To solve the optimization, the expectation in the objective is approximated using a Monte Carlo sampling scheme ~\cite{metropolis1953}. At each iteration, $M$ video sequences are rendered under diverse scene parameters $\{\mathcal{C}, \mathcal{O}, \mathcal{I}, \mathcal{B}, \Phi_g\}$ using the current control-point configuration $P_g$. The gradients of the loss with respect to the control points are estimated from these samples, and $P_g$ is updated through gradient descent. 

A time-weighted sequence-level loss extended from the single-frame loss is proposed:
\begin{equation}
  L_{\text{seq}} = \sum_{t=1}^{T} w_t\,L(x_t).
  \label{eq:Lseq}
\end{equation}
The temporal weights adopt a temperature-controlled softmax, 
\begin{equation}
  w_t = \frac{\exp\!\big(\gamma\,L(x_t)\big)}
              {\sum_{k=1}^{T} \exp\!\big(\gamma\,L(x_k)\big)}.
\end{equation}
 where $\gamma$ is the temperature coefficient. $w_t$ is used as an external weight excluded from backpropagation. To prevent control points from over-clustering during updates, a differentiable repulsive control loss~\cite{hu2023advcamocloth} (ctrlLoss) in UV space is introduced. This loss penalizes pairwise distances between control points, thereby suppressing near neighbors, promoting spatial dispersion and uniform coverage, while preserving printability. The normalization term ensures that the loss remains close to zero under approximately uniform layouts.
\begin{equation}
  L_{\text{ctrl}}
  =\frac{1}{N^2}\sum_{i,j}\exp\!\big(-d_{ij}^2/\sigma^2\big)\; -\; \frac{1}{N},
\end{equation}
where $N$ denotes the number of control points and $\sigma$ controls the radius of repulsion. When control points cluster, $\exp(-d_{ij}^2/\sigma^2)$ increases, thereby raising $L_{\text{ctrl}}$ to penalize excessive clustering and encourage spatial dispersion.

Finally, the iterative process gradually refines the control points to minimize $L_{\text{iter}}=L_{\text{seq}}\; +\; \lambda\cdot L_{\text{ctrl}}$ across varying camera, motion, illumination, and background conditions, where $\lambda$ denotes the regularization weight, balancing temporal robustness and printability.



\section{Experiments}
\label{sec:experiments}

\subsection{Settings}
All experiments are conducted on an NVIDIA RTX 5090 GPU. Due to page limitations, only the key information is described, and the full discussion on parameters and protocols are provided in the Appendix.

\subsubsection{Training Details}
\vspace{-3mm}
\begin{figure}[H]
  \centering
  \begin{minipage}[t]{0.18\linewidth}
    \centering
    \vspace{0pt}%
    \includegraphics[width=\linewidth]{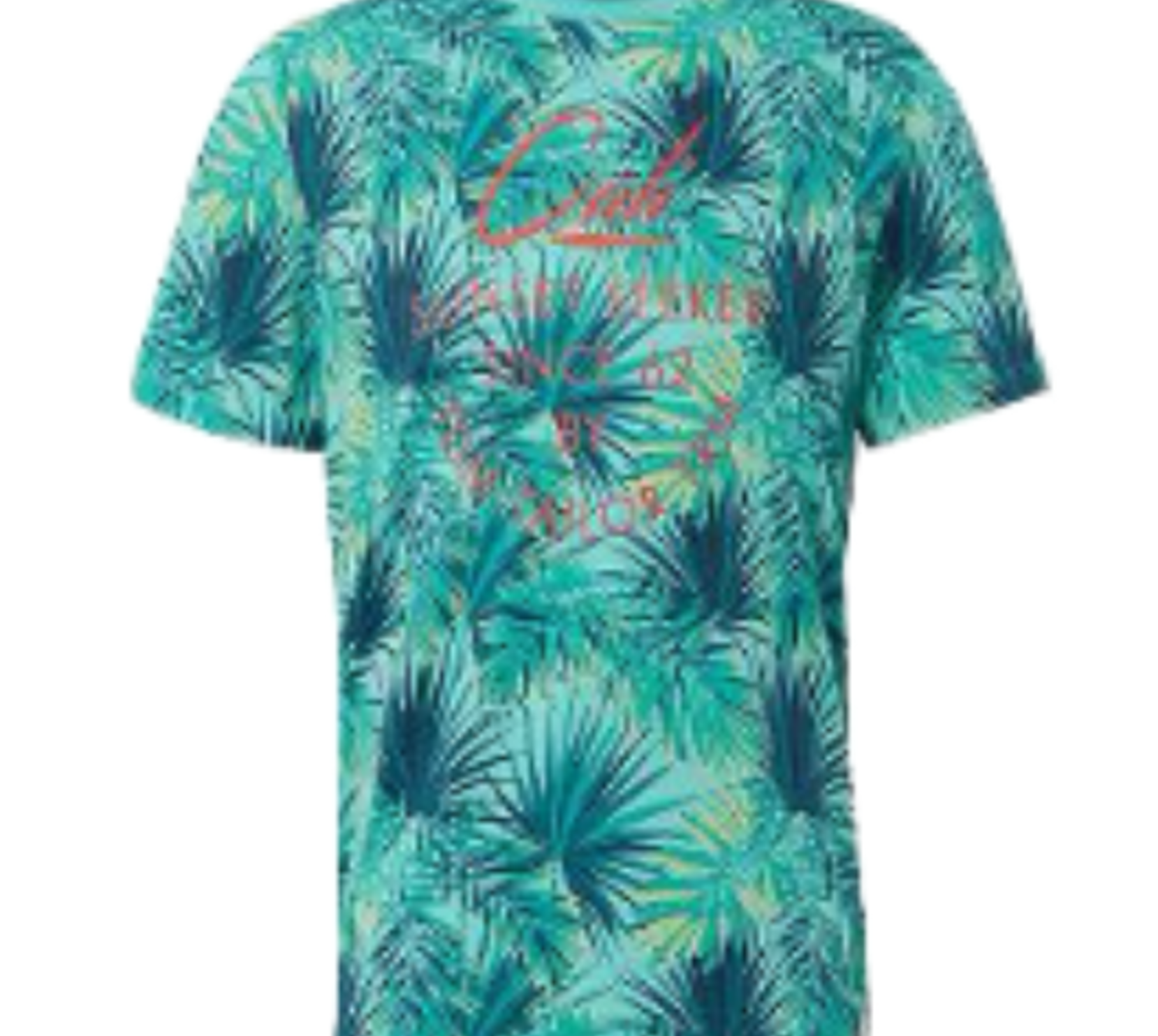}
    \vspace{2pt}
    {\scriptsize\strut $I_\mathrm{upper}$ \strut}
  \end{minipage}
  \hspace{0.04\linewidth}
  \begin{minipage}[t]{0.18\linewidth}
    \centering
    \vspace{0pt}%
    \includegraphics[width=\linewidth]{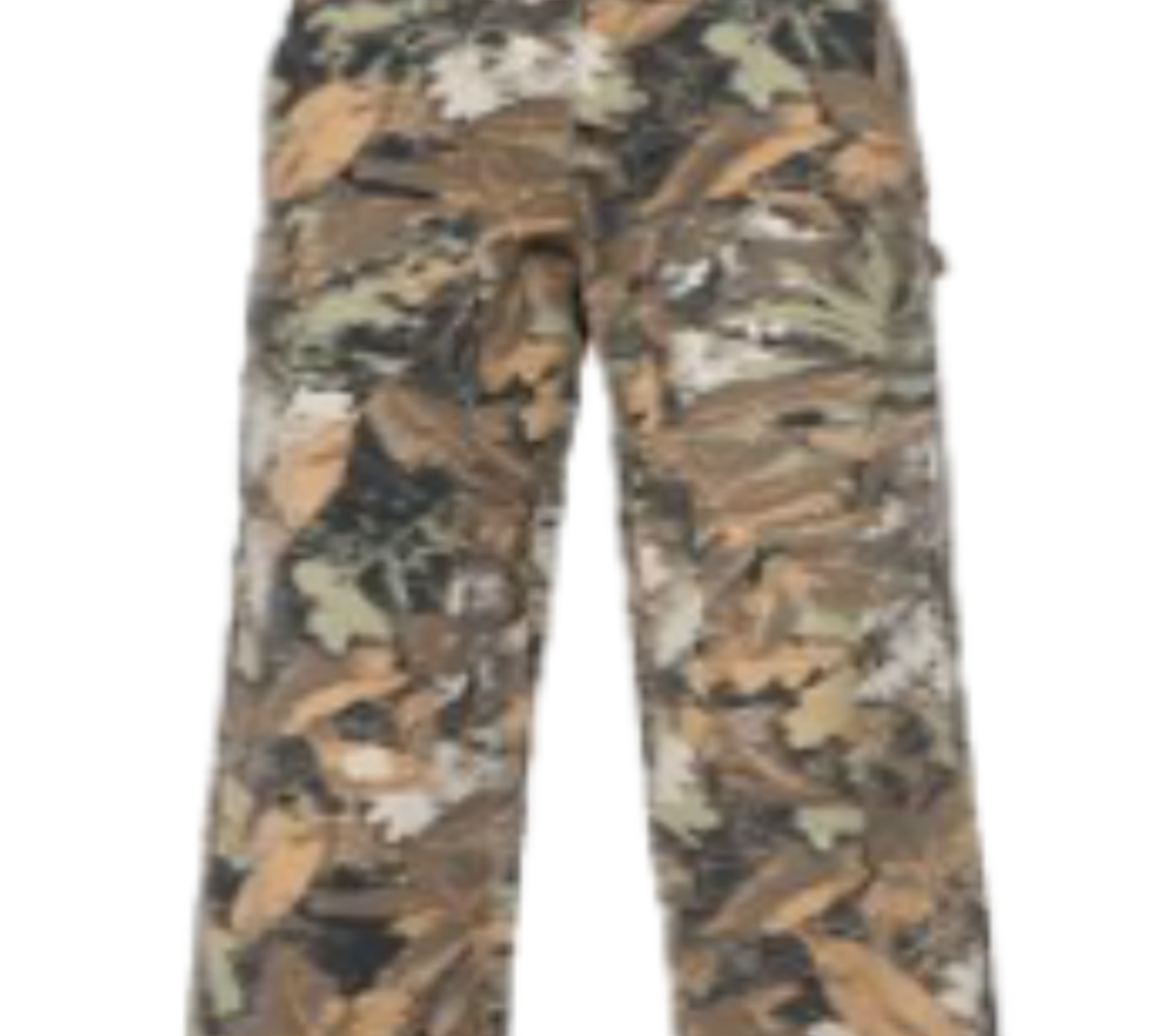}
    \vspace{2pt}
    {\scriptsize\strut $I_\mathrm{lower}$\strut}
  \end{minipage}
  \hspace{0.04\linewidth}
  \begin{minipage}[t]{0.18\linewidth}
    \centering
    \vspace{0pt}%
    \includegraphics[width=\linewidth]{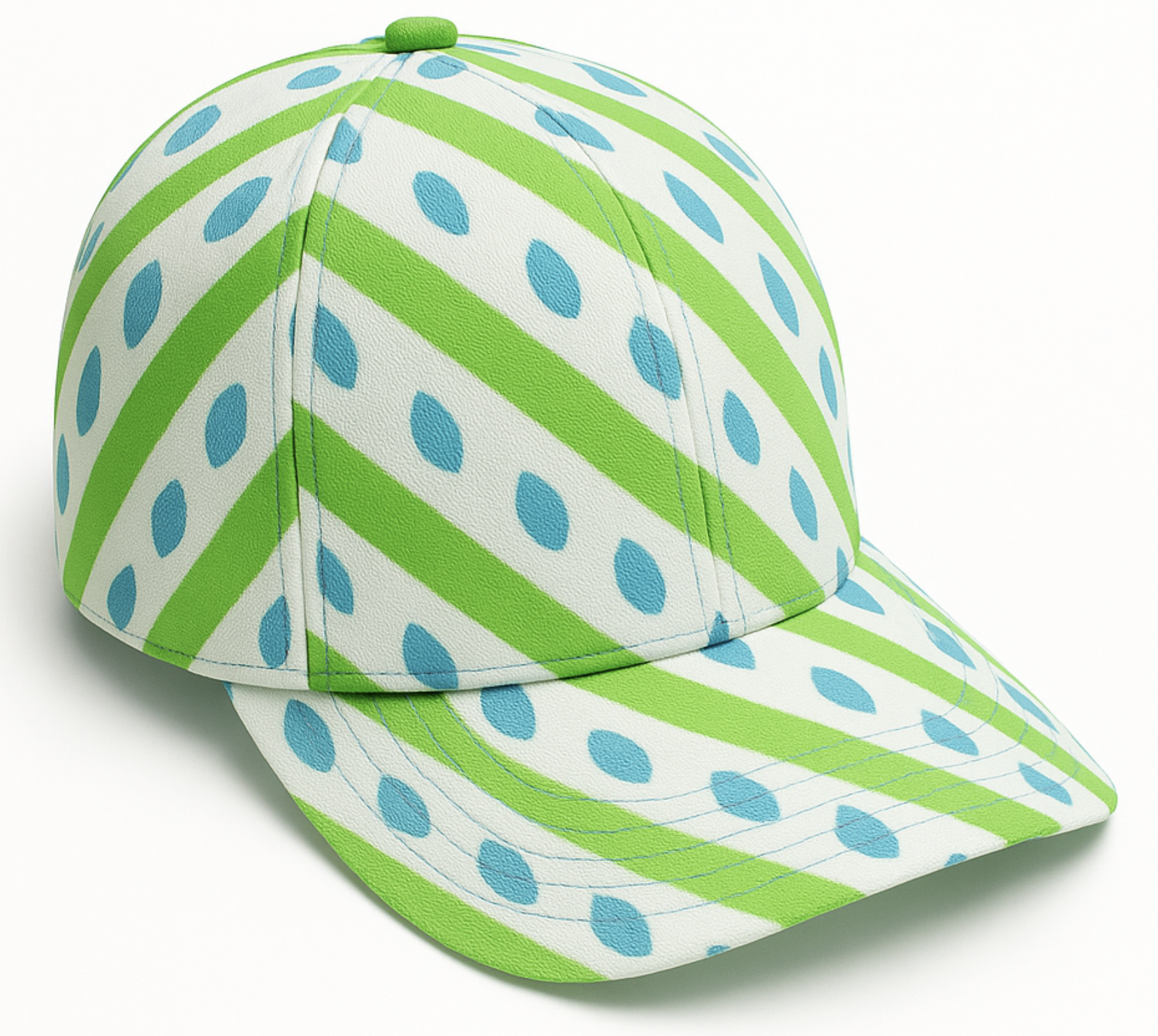}
    \vspace{2pt}
    {\scriptsize\strut $I_\mathrm{hat}$\strut}
  \end{minipage}
  \vspace{-3mm}
  \caption{The product images used in our model.}
  \label{fig:figures of product}
\end{figure}
\vspace{-3mm}
\noindent\textbf{Texture Initialization:} Southeast Asian-style garments are selected as product images $I_\mathrm{upper}$, $I_\mathrm{lower}$, and $I_\mathrm{hat}$ (Fig.~\ref{fig:figures of product}) since, even after texture modification, their overall patterns (Fig.~\ref{fig:Ablation study in physical}) remain realistic and visually plausible. $K = 6$ and $P_{\max} = 600$ are used to balance spatial fidelity and computational efficiency.

\noindent\textbf{Sequence Generation:}  Texture generation employs the variable-type seed mechanism with a mixing ratio of $0.7$ and a clamp shift of $0.01$. Color mixing follows the Gumbel–Softmax~\cite{Jang2017GumbelSoftmax,maddison2017the} formulation ($\tau = 0.3$) with a color-prior blur of $1$ and a regularization weight of $10$. The subject walks forward at approximately 1 m/s along the horizontal axis from an initial backward offset of 1 m, introducing consistent motion across sequences. Each sequence consists of $12$ interpolated key poses and $4$ additional static hold frames at the end to maintain temporal continuity during rendering. Camera viewpoints are uniformly sampled with elevation $e \in [40,70]^{\circ}$ and azimuth $a \in [0,360]^{\circ}$, while the camera–subject distance is fixed at 4m. The MegaDepth~\cite{Li_2018_CVPR} and ZInD~\cite{Cruz_2021_CVPR} datasets are used as background sources and are split evenly into training and validation sets. To increase physical diversity, cloth material parameters are randomized in each episode, covering a broad spectrum of realistic behaviors from soft and lightweight to stiff and heavy. The rendering resolution is fixed at $416\times416$.

\noindent\textbf{Optimization:} YOLOv3 \cite{yolov3} is used as the victim detector in training. We use the Adam optimizer~\cite{kingma2015adam} with learning rates $0.01$ for texture generation and $0.001$ for control-point parameters. Training is performed for 1,000 epochs with a batch size of 8. The temperature of softmax weighting is $\gamma = 2.0$.

\noindent\textbf{Physical Attack:} 
The adversarial texture $T_g^P$ is directly transferred onto sublimation paper as its colors already fall within the printable ICC gamut. For the upper and lower garments, the texture is transferred from the paper onto polyester fabric via heat-press sublimation, after which the fabric is cut and sewn into the final clothing pieces. For the hat, its texture is directly transferred onto a blank white hat using the same heat-press process. This workflow ensures that the printed patterns remain color-accurate, stable, and consistent with the digital textures.

\subsubsection{Evaluation Settings}
\noindent\textbf{Test Set:} For all digital experiments, test sequences are generated by combining the generated garment texture with material properties, multiple camera viewpoints, scene conditions, and background images. Background images are randomly drawn from the test pool and a video sequence is rendered, producing a diverse set of viewpoint–background combinations. For physical-world evaluation, real videos are recorded under multiple camera elevations and azimuths, with both indoor and outdoor scenes. All detectors are tested on the same collection of recorded videos with their default resolution. 

\noindent\textbf{Metrics:} Three sequence-level metrics are used to evaluate attack success and temporal stability. \emph{The sequence-level attack success rate} (\textbf{SeqASR}) is extended from the conventional image-level attack success rate~\cite{hu2023advcamocloth}. For each video, the proportion of frames satisfying $\mathrm{conf}<\tau$ or $\mathrm{IoU}<\tau_{\mathrm{IoU}}$ is computed, and the overall SeqASR is obtained by averaging these values across videos. To assess worst-case exposure, \emph{the Conditional Value-at-Risk} (\textbf{CVaR}) ~\cite{RockafellarUryasev2000} is adopted. For each video, the mean confidence of the upper $\alpha$ tail (after IoU gating) is calculated as $\mathrm{CVaR}$, and the dataset-level value is the average across videos. A higher $\mathrm{CVaR}$ indicates stronger residual detectability (weaker attack). A batch-level metric, \emph{Non-Detection Rate} (\textbf{NDR}), is introduced. NDR measures the proportion of videos where $\max_{t}\,\mathrm{conf}_v[t] < \tau$ and $\max_{t}\,\mathrm{IoU}_v[t] < \tau_{\mathrm{IoU}}$ (where $\mathrm{conf}_v[t]$ is the detector confidence on frame $t$), meaning that even the most detectable frame in each sequence fails to trigger correct recognition, thus representing complete video-level concealment. Unless otherwise specified, we set the confidence threshold to $\tau = 0.3$, the IoU threshold to $\tau_{\mathrm{IoU}} = 0.1$, and the tail parameter to $\alpha = 0.1$ throughout this paper.

\subsection{Results and Discussion}

\subsubsection{Comparison with State-of-the-Art Methods}

This section compares the proposed method against four recent open-source state-of-the-art human-detection evasion approaches: AdvGAN~\cite{AdvGAN}, AdvTexture~\cite{hu2022advtexture}, AdvCaT~\cite{hu2023advcamocloth}, and FnFAttack~\cite{FandF}. Official implementations are used, and all configurations follow the authors’ original settings to ensure fairness. For digital evaluation, 2160 synthesized videos are generated to span a broad range of scene variations. In addition, a physical-world evaluation is performed for our method using 200 video clips. 

\noindent\textbf{Overall Performance:} Table~\ref{tab:general-performance} reports the overall attack performance on the YOLOv3~\cite{yolov3} human-detection system, evaluated using three complementary sequence-level metrics: SeqASR, CVaR~\cite{RockafellarUryasev2000}, and NDR. The proposed method achieves the highest SeqASR of 94.7\% and the lowest CVaR of 22.0, indicating both superior average and worst-case concealment effectiveness. 
The NDR of 73.6\% further shows that our method successfully maintains invisibility across the entire video in most sequences, demonstrating strong temporal consistency and robustness. When physical rendering constraints are introduced (Ours (Physical)), the attack remains highly effective (SeqASR 86.2\%), validating that the generated adversarial textures are physically realizable while preserving strong performance. In contrast, all prior attack methods exhibit significantly lower SeqASR and higher CVaR, indicating limited robustness to temporal variation and camera motion. FnFAttack~\cite{FandF} shows the weakest performance, confirming that erasing detections or adding brief spurious boxes cannot replace consistent, sequence-wide suppression.

\begin{table}[t]
  \centering
  \caption{Overall performance of the proposed method compared with recent state-of-the-art evasion attacks. Results show that our approach achieves the highest SeqASR, lowest CVaR, and highest NDR, demonstrating strong and stable sequence-level evasion.}
  \vspace{-2mm}
  \label{tab:general-performance}

  \begingroup
  \fontsize{7}{9}\selectfont
  \setlength{\tabcolsep}{2.2pt}
  \renewcommand{\arraystretch}{0.88}
  \setlength{\aboverulesep}{0.1ex}
  \setlength{\belowrulesep}{0.2ex}

  \resizebox{\linewidth}{!}{%
  \begin{tabular}{@{}lccc@{}}
    \toprule
    Method & SeqASR ($\uparrow$) & CVaR ($\downarrow$) & NDR ($\uparrow$) \\
    \midrule
    AdvGAN (ICCV'21)       & 40.9 $\pm$ 33.7 & 85.3 $\pm$ 24.2 &  4.2 $\pm$  5.7 \\
    AdvTexture (CVPR'22)   & 80.7 $\pm$ 25.2 & 48.5 $\pm$ 37.3 & 36.8 $\pm$ 13.8 \\
    AdvCaT (CVPR'23)       & 40.8 $\pm$ 33.4 & 86.5 $\pm$ 23.3 &  4.2 $\pm$  2.4 \\
    FnFAttack (ICCV'23)    & 28.6 $\pm$ 31.3 & 91.2 $\pm$ 19.3 &  2.8 $\pm$  4.8 \\
    \addlinespace[1pt]
    \textbf{Ours}            & \textbf{94.7 $\pm$ 11.4} & \textbf{22.0 $\pm$ 31.6} & \textbf{73.6 $\pm$ 10.7} \\
    \textbf{Ours (Physical)} & \textbf{86.2 $\pm$ 9.5}  & \textbf{51.6 $\pm$ 27.0} & \textbf{39.6 $\pm$ 8.2} \\
    \bottomrule
  \end{tabular}
  }

  \endgroup
  \vspace{-2mm}
\end{table}

\noindent\textbf{Transferability on Detection Models:} This section evaluates the transferability of each attack method by testing them on five human-detection models: YOLOv3~\cite{yolov3}, YOLOv8~\cite{yolov8}, YOLOX~\cite{yolox2021}, SSD~\cite{SSD}, and Deformable DETR~\cite{DDETR}. All methods are optimized on YOLOv3~\cite{yolov3}, except FnFAttack~\cite{FandF} which is trained using YOLOX~\cite{yolox2021}. The resulting SeqASR scores across these detectors are reported in Table~\ref{tab:transferability}.

Our approach achieves consistently high transferability, with SeqASR above 84\% on all detectors when digitally rendered, including both anchor-based (YOLOv3~\cite{yolov3}, YOLOv8~\cite{yolov8}) and anchor-free architectures (YOLOX~\cite{yolox2021}, DDETR~\cite{DDETR}). This stability indicates that the learned adversarial pattern captures model-agnostic vulnerabilities by optimizing over long video sequences, realistic cloth dynamics, and diverse scene transformations. The physical results further demonstrate strong generalization, achieving SeqASR above 60–86\% despite real-world imperfections such as fabric deformation, printing shifts, illumination variation, and camera noise. These findings confirm that sequence-aware, physically grounded optimization produces adversarial textures that transfer reliably across architectures, far beyond what can be achieved with traditional per-frame or purely digital methods. However, all baselines show limited cross-model robustness. AdvGAN~\cite{AdvGAN}, AdvCaT~\cite{hu2023advcamocloth} and FnFAttack~\cite{FandF} perform poorly across all detectors, with SeqASR values mostly below 50\%, revealing severe overfitting to the training architecture. Their perturbations fail to remain effective once the detector’s feature extractor or prediction head changes. AdvTexture~\cite{hu2022advtexture} achieves stronger transferability, particularly on SSD~\cite{SSD} and DDETR~\cite{DDETR}, but still exhibits high variance and inconsistent performance. Its frame-wise optimization makes it sensitive to detector-specific biases, causing effectiveness to drop substantially on YOLO-based models.

\begin{table}[t]
  \centering
  \caption{SeqASR transferability across five human-detection models. All attacks are optimized on YOLOv3 (except FnFAttack, which uses YOLOX) and evaluated on YOLOv3, YOLOv8, SSD, DDETR, and YOLOX to assess cross-model generalization.}
  \vspace{-2mm}
  \label{tab:transferability}

  \begingroup
  \fontsize{7}{9}\selectfont
  \setlength{\tabcolsep}{2.2pt}
  \renewcommand{\arraystretch}{0.8}
  \setlength{\aboverulesep}{0.1ex}
  \setlength{\belowrulesep}{0.2ex}

  \resizebox{\linewidth}{!}{%
  \begin{tabular}{@{}lccccc@{}}
    \toprule
     & YOLOv3 & YOLOv8 & YOLOX & SSD & DDETR \\
    \midrule
    AdvGAN          & 41.2 $\pm$ 33.2 & 37.6 $\pm$ 31.3 & 34.7 $\pm$ 33.3 & 33.2 $\pm$ 32.0 & 39.7 $\pm$ 32.3 \\
    AdvTexture      & 80.9 $\pm$ 25.2 & 55.0 $\pm$ 32.1 & 71.5 $\pm$ 29.0 & 84.0 $\pm$ 21.4 & 86.3 $\pm$ 18.5 \\
    AdvCaT          & 41.3 $\pm$ 33.4 & 40.1 $\pm$ 31.1 & 48.2 $\pm$ 35.1 & 29.4 $\pm$ 30.4 & 18.0 $\pm$ 24.4 \\
    FnFAttack       & 29.3 $\pm$ 31.4 & 27.2 $\pm$ 32.3 & 40.5 $\pm$ 35.7 & 26.9 $\pm$ 30.9 & 29.8 $\pm$ 31.4 \\
    \addlinespace[1pt]
    \textbf{Ours}            & \textbf{94.7 $\pm$ 11.4} & \textbf{95.0 $\pm$ 10.8} & \textbf{84.8 $\pm$ 22.2} & \textbf{87.7 $\pm$ 16.4} & \textbf{91.1 $\pm$ 14.8} \\
    \textbf{Ours (Physical)} & \textbf{86.2 $\pm$ 9.5}  & \textbf{84.1 $\pm$ 18.2} & \textbf{80.9 $\pm$ 12.0} & \textbf{69.6 $\pm$ 21.2} & \textbf{62.8 $\pm$ 22.8} \\
    \bottomrule
  \end{tabular}
  }

  \endgroup
\end{table}

\noindent\textbf{Transferability on Camera Elevation Angles:} Camera elevation is an important factor because human-detection models often exhibit significant performance variation under changes in viewpoint. Therefore, a strong adversarial attack must remain effective across these geometric shifts. Table~\ref{tab:angle-scenarios} reports the SeqASR results at four elevation angles (40°, 50°, 60°, 70°). Only digital attacks are evaluated here, as physical-world settings cannot precisely control the camera elevation.

Across all elevation angles, our method achieves the highest SeqASR scores with low variance, demonstrating consistent and stable evasion performance. Performance is strong even at extreme elevation conditions, maintaining SeqASR above 87\%, and reaching above 95\% for elevations of 50–70°, indicating that the adversarial pattern generalizes well across perspective changes. In contrast, existing approaches degrade rapidly and exhibit strong instability. AdvGAN~\cite{AdvGAN} and FnFAttack~\cite{FandF} perform particularly poorly at lower elevations, with SeqASR dropping below 25\% at 40°. AdvCaT~\cite{hu2023advcamocloth}, although effective at near eye-level viewpoints reported in its original setting, fails almost entirely under all elevated viewpoints in this evaluation, highlighting the sensitivity of its frame-wise optimization to geometric changes. AdvTexture~\cite{hu2022advtexture} retains moderate robustness and performs better at higher angles, yet its variance remains large and its performance at lower angles remains unreliable. 

\begin{table}[H]
  \centering
  \caption{SeqASR performance of different attack methods across varying camera elevation angles. Higher SeqASR indicates stronger viewpoint-robust evasion.}
  \vspace{-2mm}
  \label{tab:angle-scenarios}

  \begingroup
  \fontsize{7}{9}\selectfont
  \setlength{\tabcolsep}{2.2pt}
  \renewcommand{\arraystretch}{0.88}
  \setlength{\aboverulesep}{0.1ex}
  \setlength{\belowrulesep}{0.2ex}
  \resizebox{\linewidth}{!}{%
  \begin{tabular}{@{}lcccc@{}}
    \toprule
            & $40^\circ$           & $50^\circ$           & $60^\circ$           & $70^\circ$           \\
    \midrule
    AdvGAN (ICCV'21)     & 22.5 $\pm$ 26.7 & 38.5 $\pm$ 33.7 & 43.3 $\pm$ 32.6 & 59.4 $\pm$ 30.4 \\
    AdvTexture (CVPR'22) & 67.1 $\pm$ 29.1 & 80.6 $\pm$ 27.1 & 83.5 $\pm$ 23.2 & 91.7 $\pm$ 10.2 \\
    AdvCaT (CVPR'23)     & 21.0 $\pm$ 24.4 & 40.3 $\pm$ 36.8 & 43.8 $\pm$ 29.5 & 58.2 $\pm$ 30.6 \\
    FnFAttack (ICCV'23)  &  7.3 $\pm$ 15.6 & 22.6 $\pm$ 27.4 & 31.8 $\pm$ 28.5 & 52.5 $\pm$ 32.1 \\
    \addlinespace[1pt]
    \textbf{Ours}        & \textbf{87.2 $\pm$ 18.4} & \textbf{97.3 $\pm$ 6.2} & \textbf{96.9 $\pm$ 6.4} & \textbf{95.7 $\pm$ 7.2} \\
    \bottomrule
  \end{tabular}
  }
  \endgroup
  \vspace{-2mm}
\end{table}

\vspace{-4mm}

\begin{table}[H]
  \centering
  \caption{SeqASR performance under different garment-material presets. Denim, cotton, and silk/chiffon T-shirts are tested to cover a wide range of stiffness and density conditions. The proposed method maintains high attack success across all materials, while baseline methods show noticeable sensitivity to material changes.}
  \vspace{-2mm}
  \label{tab:materials-scenarios}

  \begingroup
  \fontsize{7}{9}\selectfont
  \setlength{\tabcolsep}{2.2pt}
  \renewcommand{\arraystretch}{0.88}
  \setlength{\aboverulesep}{0.1ex}
  \setlength{\belowrulesep}{0.2ex}

  \resizebox{\linewidth}{!}{%
  \begin{tabular}{@{}lccc@{}}
    \toprule
            & Denim T-shirt       & Cotton T-shirt      & Chiffon T-shirt     \\
    \midrule
    AdvGAN (ICCV'21)     & 41.3 $\pm$ 33.9 & 40.3 $\pm$ 34.0 & 37.2 $\pm$ 32.3 \\
    AdvTexture (CVPR'22) & 79.9 $\pm$ 26.1 & 80.1 $\pm$ 26.4 & 83.7 $\pm$ 24.0 \\
    AdvCaT (CVPR'23)     & 38.6 $\pm$ 33.1 & 39.3 $\pm$ 33.6 & 39.7 $\pm$ 32.3 \\
    FnFAttack (ICCV'23)  & 28.7 $\pm$ 31.6 & 28.3 $\pm$ 31.3 & 26.5 $\pm$ 29.8 \\
    \addlinespace[1pt]
    \textbf{Ours}        & \textbf{91.0 $\pm$ 13.1} & \textbf{90.1 $\pm$ 14.3} & \textbf{92.2 $\pm$ 12.1} \\
    \bottomrule
  \end{tabular}
  }

  \endgroup
  \vspace{-2mm}
\end{table}
\noindent\textbf{Transferability on Garment Material:} The influence of garment material on attack robustness is evaluated using three representative fabric presets, shown in Fig.~\ref{fig:tshirt-row}. These presets cover a broad spectrum of physical behaviors, ranging from heavy and stiff (denim), to moderately flexible (cotton), to lightweight and highly drapeable (silk/chiffon). Only digital attacks are considered in this experiment.

As shown in Table ~\ref{tab:materials-scenarios}, substantial degradation is consistently observed in all prior methods when the material properties change. The frame-based attack approaches yield SeqASR values around 25-40\% across all materials, indicating that their perturbations are tightly coupled to the cloth geometry seen during training and are easily disrupted by changes in wrinkle patterns, flutter dynamics, and silhouette deformation. AdvTexture~\cite{hu2022advtexture} provides higher performance (approximately 80\% SeqASR), but its results still vary noticeably across materials, suggesting limited robustness to dynamic shape variations introduced by soft or stiff fabrics. In contrast, the proposed method maintains strong and stable performance across all three material types, achieving SeqASR scores above 90\% with low variance. This consistency indicates that the learned perturbations are not reliant on any single deformation pattern, but instead remain effective under a wide range of cloth behaviors—from the rigid folds of denim to the high-frequency flutter exhibited by silk chiffon. The robustness is attributed to the integration of sequence-level EOT~\cite{EOT} and physics-aware garment simulation, which forces the optimization to account for temporal variations in cloth motion and viewpoint-dependent geometry. 

\noindent\textbf{Attack Stability Under Different IoU Thresholds: }IoU thresholds strongly influence human-detection outcomes. A higher IoU threshold helps avoid confusion from overlapping bounding boxes, but an excessively strict threshold can overestimate attack success, whereas a very loose threshold introduces undesirable noise. Table~\ref{tab:iou} reports SeqASR values under IoU thresholds ranging from extremely permissive (0.01) to strict (0.5).

Across all thresholds, the proposed method consistently achieves the highest SeqASR values, with performance rising from 87-89\% at IoU 0.01 to over 96\% at IoU 0.5. This monotonic improvement shows that the detector not only fails to classify the person but also fails to localize the body even when stricter spatial alignment is required. In contrast, all baseline methods exhibit much lower SeqASR values and almost no sensitivity to IoU changes. Their performance remains nearly flat across thresholds, indicating that detection failures occur primarily due to classification suppression rather than localization disruption. AdvTexture~\cite{hu2022advtexture} provides moderate robustness but still remains far behind the proposed method at every threshold.

\vspace{-3mm}
\begin{table}[H]
  \centering
  \caption{Performance across IoU thresholds}
  \vspace{-2mm}
  \label{tab:iou}
  \begingroup
  \fontsize{7}{9}\selectfont
  \setlength{\tabcolsep}{1.5pt}
  \renewcommand{\arraystretch}{0.88}
  \setlength{\aboverulesep}{0.1ex}
  \setlength{\belowrulesep}{0.2ex}

  \resizebox{\linewidth}{!}{%
  \begin{tabular}{@{}lcccc@{}}
    \toprule
    Method & IoU0.01 & IoU0.1 & IoU0.3 & IoU0.5 \\
    \midrule
    AdvGAN        & $38.2\,\pm\,33.5$ & $40.9\,\pm\,33.7$ & $40.7\,\pm\,33.7$ & $41.0\,\pm\,34.5$ \\
    AdvTexture    & $76.9\,\pm\,26.1$ & $80.7\,\pm\,25.4$ & $80.9\,\pm\,25.2$ & $82.3\,\pm\,24.8$ \\
    AdvCaT        & $38.8\,\pm\,33.3$ & $40.3\,\pm\,30.5$ & $40.8\,\pm\,33.4$ & $41.5\,\pm\,33.6$ \\
    FnFAttack     & $27.9\,\pm\,32.4$ & $28.6\,\pm\,31.3$ & $28.6\,\pm\,31.2$ & $28.9\,\pm\,31.9$ \\
    \textbf{Ours} & $\mathbf{87.3}\,\pm\,\mathbf{14.3}$ & $\mathbf{94.2}\,\pm\,\mathbf{11.5}$ & $\mathbf{94.3}\,\pm\,\mathbf{11.6}$ & $\mathbf{96.3}\,\pm\,\mathbf{12.1}$ \\
    \bottomrule
  \end{tabular}
  }
  \endgroup
\end{table}
\vspace{-2em}
\begin{figure}[H]
  \centering
  \begin{subfigure}{0.20\linewidth}
    \centering
    \includegraphics[width=0.75\linewidth]{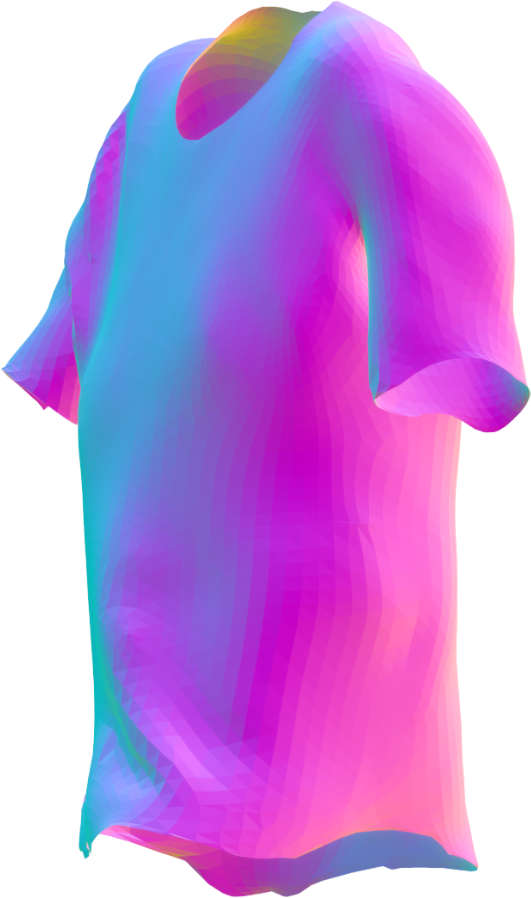}
    \subcaption{Denim}
  \end{subfigure}\hspace{0.04\linewidth}
  \begin{subfigure}{0.20\linewidth}
    \centering
    \includegraphics[width=0.75\linewidth]{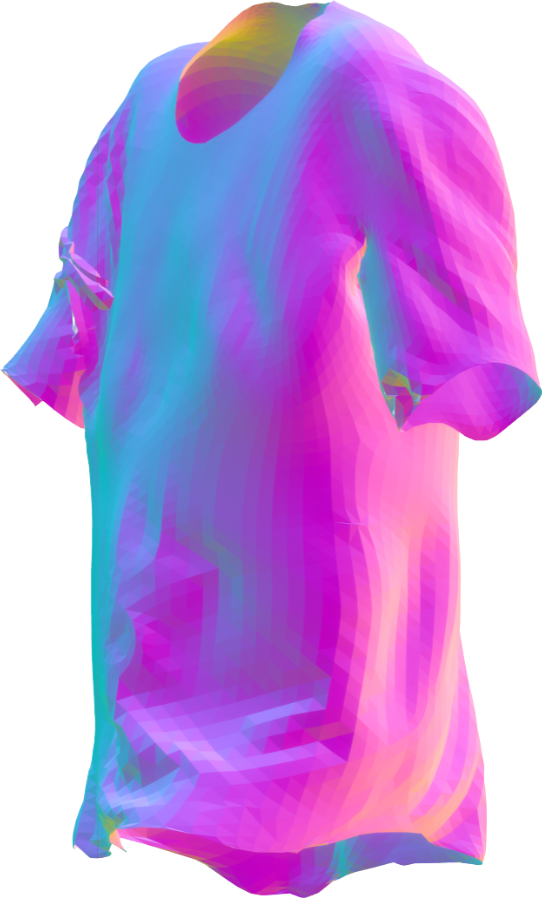}
    \subcaption{Cotton}
  \end{subfigure}\hspace{0.04\linewidth}
  \begin{subfigure}{0.20\linewidth}
    \centering
    \includegraphics[width=0.75\linewidth]{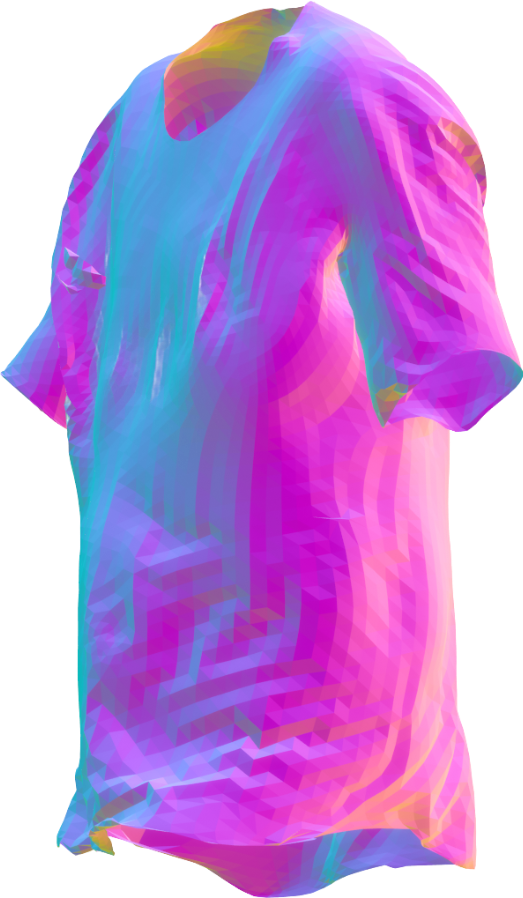}
    \subcaption{Chiffon}
  \end{subfigure}
  \caption{Visualization of the three garment-material presets used in the material-robustness evaluation.}
  \label{fig:tshirt-row}
\end{figure}

\begin{figure}[H]
  \centering
  \hspace*{-10mm}
  \includegraphics[width=\linewidth]{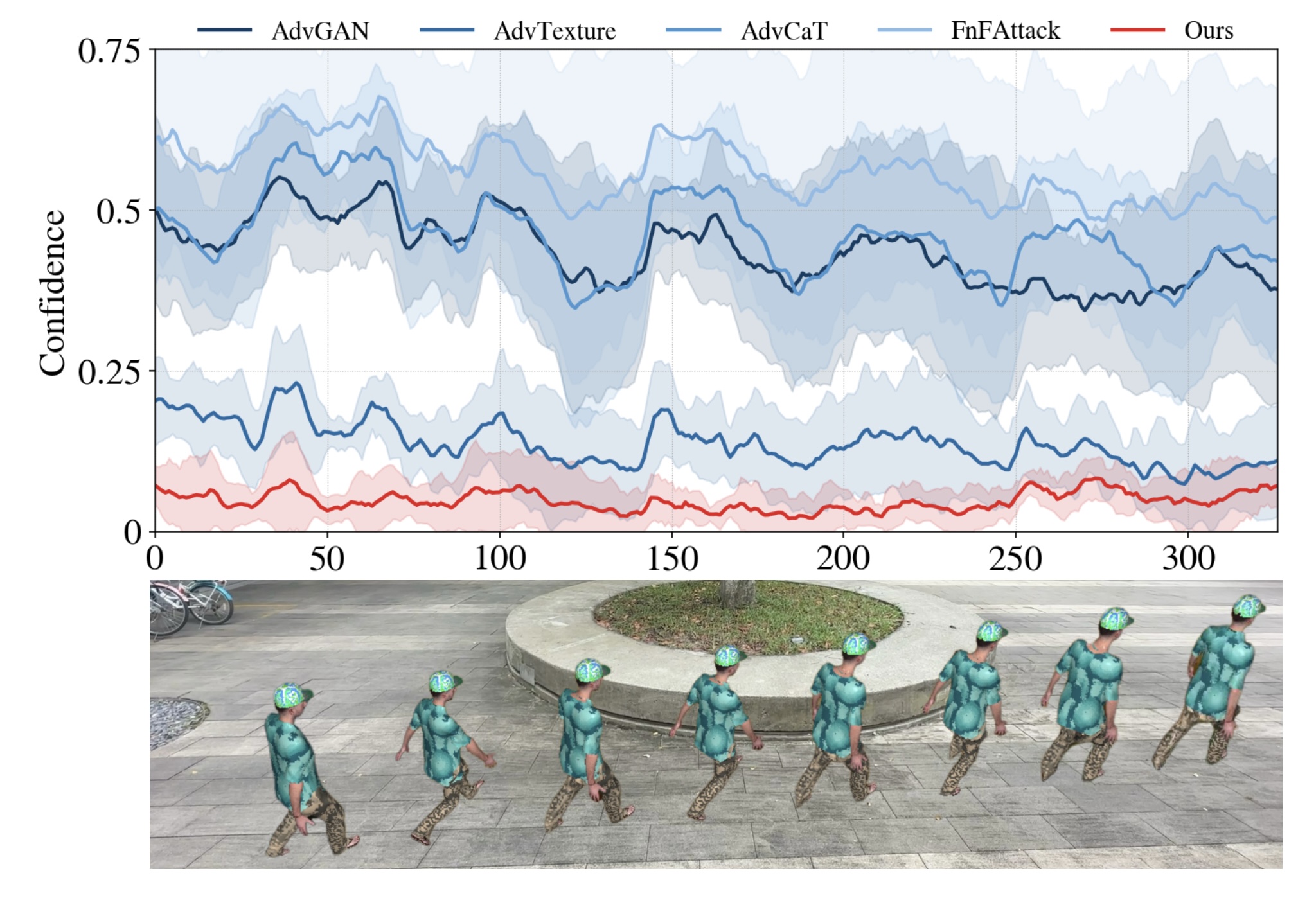}
  \caption{Confidence evolution over a 327-frame walking sequence. Our method sustains near-zero, stable confidence throughout the motion, whereas baselines fluctuate heavily under dynamic pose, deformation, and viewpoint changes.}
  \label{fig:curves-repeat3}
  \vspace{-2mm}
\end{figure}

\vspace{-5mm}
\noindent\textbf{Attack Stability in Long Video Sequences:} A visualized example is provided to examine how detection confidence evolves over an extended walking sequence of 327 frames, highlighting the temporal robustness of different attack methods. The sequence and its corresponding confidence trajectories are shown in Fig.~\ref{fig:curves-repeat3}.

Across the full 327-frame sequence, the proposed method sustains consistently low detection confidence, remaining near zero with only minor variation. This indicates that the optimized texture retains its effectiveness under substantial changes in body pose, garment deformation, and camera–subject geometry. The narrow confidence band further shows that suppression of the detector is uniform over time, not dependent on momentary viewpoint alignment. However, all baseline methods exhibit high and unstable 
confidence trajectories. Although AdvTexture~\cite{hu2022advtexture} achieves relatively strong average suppression compared with other baselines, its confidence curve exhibits very large variance, frequently oscillating between low and high values as the sequence progresses. This instability reflects vulnerability to changes in pose, cloth dynamics, and viewpoint. Other baselines fluctuate even more severely, with repeated confidence spikes that indicate frequent breakdowns in adversarial effect. 

\vspace{-3mm}
\subsubsection{Ablation Study} 

\begin{figure}[H]
  \centering
  \begin{subfigure}{0.18\linewidth}
    \centering
    \includegraphics[width=\linewidth]{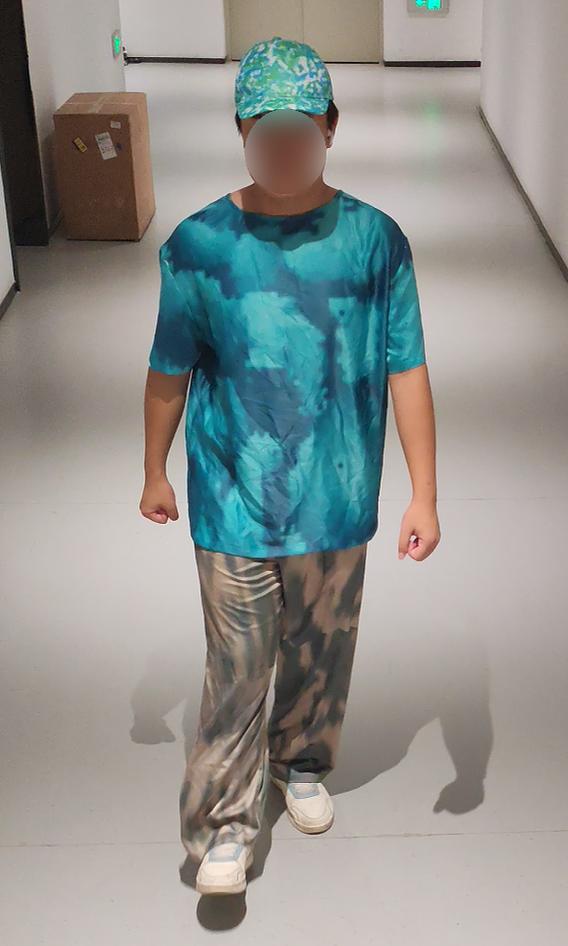}
    \subcaption{Ours}
  \end{subfigure}\hspace{0.01\linewidth}
  \begin{subfigure}{0.18\linewidth}
    \centering
    \includegraphics[width=\linewidth]{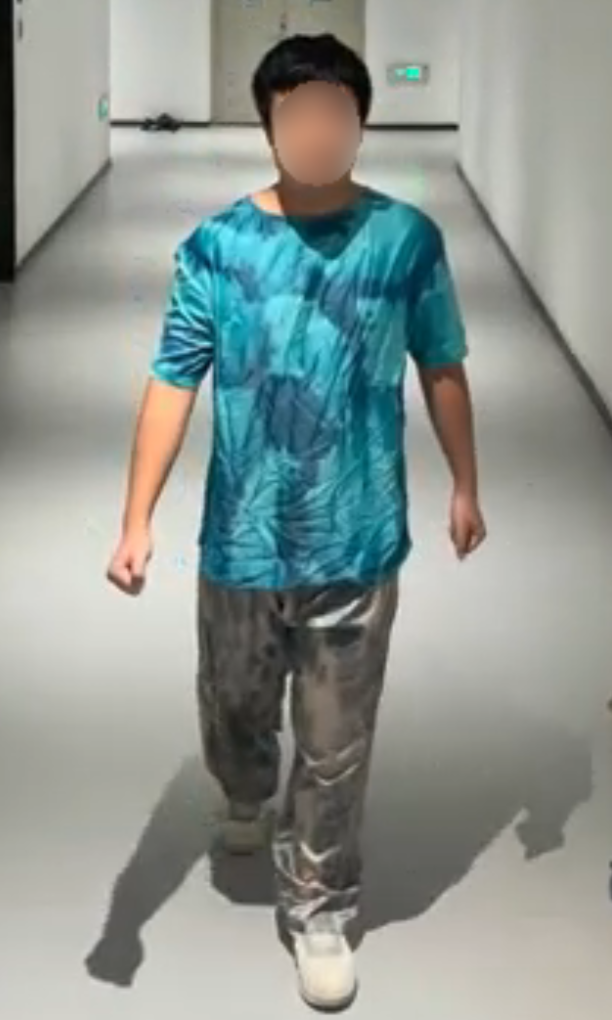}
    \subcaption{NoHat}
  \end{subfigure}\hspace{0.01\linewidth}
  \begin{subfigure}{0.18\linewidth}
    \centering
    \includegraphics[width=\linewidth]{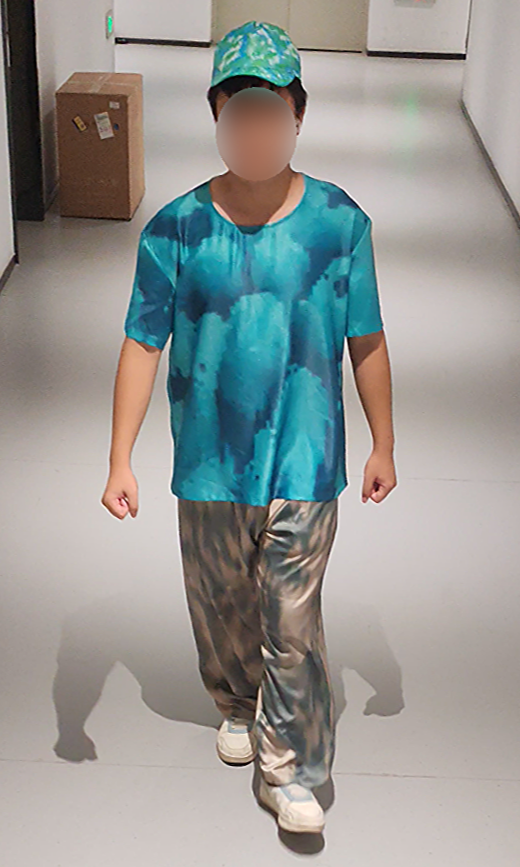}
    \subcaption{NoSeq}
  \end{subfigure}\hspace{0.01\linewidth}
  \begin{subfigure}{0.18\linewidth}
    \centering
    \includegraphics[width=\linewidth]{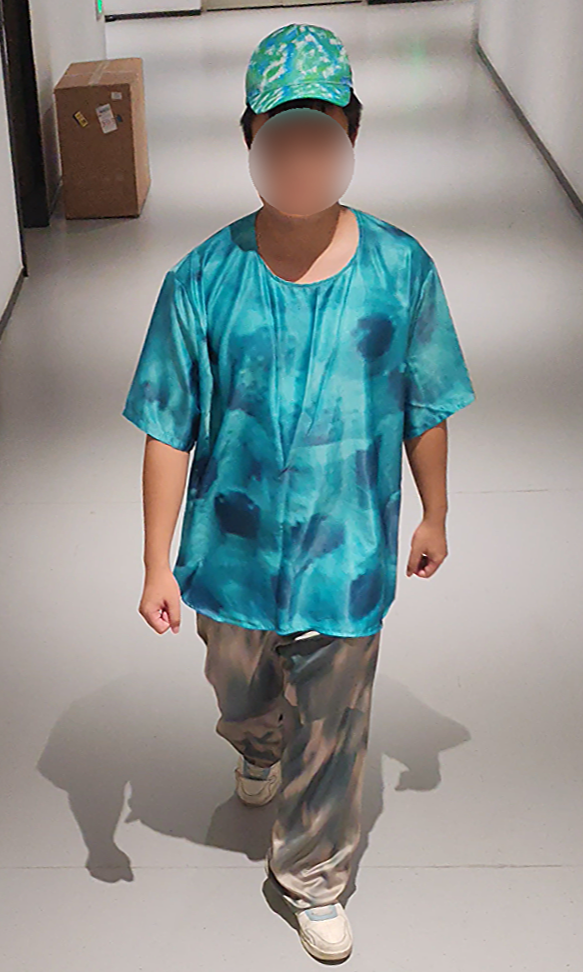}
    \subcaption{NoHood}
  \end{subfigure}
\vspace{-2mm}
  \caption{Ablation study in physical deployment.}
  \label{fig:Ablation study in physical}
  \vspace{-2mm}
\end{figure}

The contribution of major components of the proposed method, including hat texture, sequence-level optimization, and HOOD-based physical cloth simulation, are evaluated under both digital and physical attack settings (Fig.~\ref{fig:Ablation study in physical}), corresponding to NoHat, NoSeq, NoHood in Table ~\ref{tab:ablation}.

Across all metrics, the full model (Ours) achieves the strongest performance, confirming that its components work in a complementary manner. Removing individual modules consistently degrades SeqASR, worsens worst-case exposure (CVaR increases), and reduces the proportion of sequences where the detector fails entirely (NDR decreases). The results highlight that each design choice contributes meaningfully to robustness.

The hat component plays a surprisingly important role. When the hat is removed (NoHat), SeqASR drops sharply from 94.7 to 70.9 (digital) and from 86.2 to 61.7 (physical). NDR also drops drastically, showing that attacks become much less reliable across all frames. This effect is likely due to the head region being a dominant cue for human presence and localization, meaning that suppressing detector responses near the head significantly strengthens sequence-level evasion stability.

Without sequence-level optimization (NoSeq), digital SeqASR decreases to 86.1 and physical SeqASR to 71.3. More critically, NDR collapses to 4.2 in the physical setting, indicating failure to maintain suppression throughout entire sequences. This demonstrates that optimizing textures only at the frame level cannot sustain robust attacks over long motion cycles or under real-world distortions.

The HOOD proves indispensable for real-world generalization. Removing it (NoHood) results in the largest performance degradation among all ablations. Digital SeqASR falls to 58.3 and physical SeqASR to 56.6, with CVaR rising substantially. NDR in the physical world becomes nearly zero (2.1), indicating that without physically plausible cloth behavior during training, adversarial textures cannot survive real garment dynamics. This underscores the necessity of modeling realistic drape, bending, and cloth–body interactions.

\begin{table}[H]
  \centering
  \caption{Ablation study evaluating the contribution of each system component under digital and physical settings. Removing the hat (“NoHat”), disabling sequence-level optimization (“NoSeq”), or removing physical garment simulation (“NoHood”) all lead to significant performance degradation across SeqASR, CVaR, and NDR, confirming the necessity of each module.}
  \vspace{-2mm}
  \label{tab:ablation}

  \begingroup
  \fontsize{7}{9}\selectfont
  \setlength{\tabcolsep}{2.2pt}
  \renewcommand{\arraystretch}{0.88}
  \setlength{\aboverulesep}{0.1ex}
  \setlength{\belowrulesep}{0.2ex}

  \newcommand{\hdrsml}[1]{{\fontsize{6}{7}\selectfont #1}}

  \resizebox{\linewidth}{!}{%
  \begin{tabular}{@{}lcccccc@{}}
    \toprule
     & \multicolumn{2}{c}{\textbf{SeqASR ($\uparrow$)}} & \multicolumn{2}{c}{\textbf{CVaR ($\downarrow$)}} & \multicolumn{2}{c}{\textbf{NDR ($\uparrow$)}} \\
    \cmidrule(lr){2-3}\cmidrule(lr){4-5}\cmidrule(lr){6-7}
     & \hdrsml{Digital} & \hdrsml{Physical} & \hdrsml{Digital} & \hdrsml{Physical} & \hdrsml{Digital} & \hdrsml{Physical} \\
    \midrule
    Ours        & 94.7 & 86.2 & 22.0 & 51.6 & 73.6 & 39.6 \\
    NoHat       & 70.9 & 61.7 & 60.5 & 73.5 & 22.2 & 12.5 \\
    NoSeq       & 86.1 & 71.3 & 41.8 & 77.0 & 43.1 &  4.2 \\
    NoHood      & 58.3 & 56.6 & 73.9 & 84.5 & 13.9 &  2.1 \\
    \bottomrule
  \end{tabular}
  }
  \endgroup
  \vspace{-2mm}
\end{table}


\section{Conclusion}
\label{sec:conclusion}
We introduce a sequence level framework for generating physically realizable adversarial garment textures. From real product photographs, we initialize textures with ICC profile–based gamut locking, then optimize them through a unified pipeline that couples UV space parameterization, physically based human garment simulation, differentiable rendering, and expectation-over-transformation. This yields adversarial textures that keep human detector confidence low across entire video sequences, while remaining natively restricted to a printable color gamut. Extensive digital and physical experiments show that our method achieves substantially higher sequence-level robustness, lower worst case exposure, and stronger cross-detector transferability than recent state-of-the-art approaches. Its ability to maintain concealment under continuous motion, changing camera viewpoints, and diverse garment materials underscores the practical importance of sequence level optimization for real world adversarial robustness. 

\clearpage
\setcounter{page}{1}
\maketitlesupplementary

This supplementary document provides additional details that could not be included in the main paper due to space constraints. Complete training settings, texture–palette initialization procedures, sequence synthesis and garment-simulation configurations, optimization schedules, and evaluation protocols are described. Additional visualizations are also provided to illustrate garment materials, simulated dynamics, and rendered test samples. All hyperparameter ranges, dataset splits, and implementation choices referenced in the main paper are fully specified here for reproducibility. The source code is included in the supplementary archive, and a public GitHub repository will be released upon acceptance.


\section{Detailed Experimental Settings}

\subsection{Training Details} 

\noindent\textbf{Texture Initialization:} Garment images representative of East and Southeast Asian clothing styles were collected from online shopping platforms such as Zalando and Jack \& Jones. For each garment, the color palette size was set to \(K = 6\), and the number of control points per color cluster was set to 600. Due to supplementary size constraints, the ICC profile used for gamut locking is included as a PNG file in the supplementary package. When a color cluster contained fewer points than required, two existing points were uniformly sampled and their arithmetic midpoint was inserted as a new point. This process was repeated until the desired number of control points was obtained. K-Means++~\cite{kmean++} initialization was used for both palette extraction and spatial clustering. Each K-Means stage was executed for 300 iterations.

\noindent\textbf{Sequence Generation:}
A walking cycle is constructed in Blender~\cite{blender42lts} using the pose data from the original SMPL paper~\cite{SMPL:2015}. Nine keyframes are first created, after which linear interpolation is applied to generate 12 intermediate frames between every adjacent keyframe, including between the last and the first keyframe, so that the motion forms a smooth loop. This results in 108 interpolated frames. Together with one anchor keyframe, the complete cycle contains 109 frames. The total number of frames is expressed as $T = M \times H + 1$, where $M$ and $H$ denote the number of keyframe intervals and interpolated frames per interval, respectively.
 
Physical garment dynamics are simulated on the interpolated human sequence using HOOD~\cite{Grigorev_2023_CVPR}. In each training episode, the cloth material parameters are randomized according to the distributions and ranges listed in Table~\ref{tab:material-params}. This sampling covers materials ranging from light to heavy and from soft to stiff. All remaining simulation parameters follow the default HOOD configuration.

\begin{table}[t]
  \centering
  \small
  \setlength{\tabcolsep}{8pt}
  \begin{tabular}{lll}
    \toprule
    \textbf{Parameter} & \textbf{Distribution} & \textbf{Range} \\
    \midrule
    $\mu$       & $\mathrm{LogUniform}$ & $[15909,\,63636]$ \\
    $\lambda$   & $\mathrm{Uniform}$    & $[3535.41,\,93333.74]$ \\
    $\kappa_b$  & $\mathrm{LogUniform}$ & $[6.37{\times}10^{-8},\,1.31{\times}10^{-3}]$ \\
    $\rho$      & $\mathrm{Uniform}$    & $[0.0434,\,0.7]$ \\
    \bottomrule
  \end{tabular}
  \caption{Randomization ranges for cloth material parameters used during training.}
  \label{tab:material-params}
\end{table}

Rendered sequences are produced with PyTorch3D~\cite{ravi2020pytorch3d} using the physically simulated garment motion. During each episode, the subject walks forward along the horizontal axis at approximately $1\,\mathrm{m/s}$, beginning from a backward offset of $1\,\mathrm{m}$. At episode initialization, a camera elevation $e \in [40^{\circ},70^{\circ}]$ and azimuth $a \in [0^{\circ},360^{\circ}]$ are sampled and held fixed for the entire episode. The camera–subject distance is fixed at $4\,\mathrm{m}$. Background images are drawn from MegaDepth and ZInD~\cite{Li_2018_CVPR,Cruz_2021_CVPR}, which are evenly partitioned into training and validation subsets. The training pool contains 2,000 background images (1,000 per dataset). For each batch, one background is sampled randomly. Rendering resolution is fixed at $416 \times 416$.

\noindent\textbf{Optimization:}
YOLOv3~\cite{yolov3} is adopted as the victim detector during training. Optimization is performed using the Adam optimizer~\cite{kingma2015adam}, with learning rates of (0.01) for texture generation and (0.001) for control-point parameters. A piecewise decay schedule is applied in which each learning rate is reduced by half every 150 epochs. Training is conducted for (1{,}000) epochs with a batch size of (8). The temperature used in the softmax weighting of the sequence loss is set to ($\gamma = 2.0$). The overall optimization objective is defined as $L_{\text{iter}} = L_{\text{seq}} + \lambda \cdot L_{\text{ctrl}}$, where the regularization weight is fixed at ($\lambda = 50$).

\subsection{Evaluation Details}

\noindent\textbf{Digital Test Set:} 
For the General Performance evaluation and the digital ablation study, four camera elevations are used, $e \in \{40^\circ, 50^\circ, 60^\circ, 70^\circ\}$. For each elevation, the azimuth is swept from $0^\circ$ to $350^\circ$ in increments of $10^\circ$. At every elevation–azimuth pair, three backgrounds are randomly sampled from the test pool, which contains 1,000 MegaDepth images and 1,000 ZInD images. This procedure produces a total of 432 digital test videos. Each video contains $218$ frames rendered at a resolution of $416\times 416$, and all remaining rendering parameters follow the training configuration.

\begin{figure}[H]
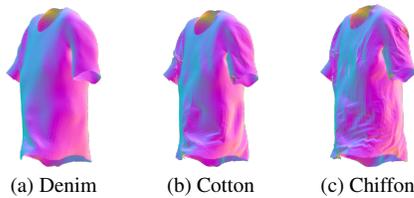

  \centering
  \begin{subfigure}{0.20\linewidth}
    \centering
    \includegraphics[width=0.75\linewidth]{pictures/Experiment/denim_T-shirt.png}
    \subcaption{Denim}
  \end{subfigure}\hspace{0.04\linewidth}
  \begin{subfigure}{0.20\linewidth}
    \centering
    \includegraphics[width=0.75\linewidth]{pictures/Experiment/cotton_T-shirt.png}
    \subcaption{Cotton}
  \end{subfigure}\hspace{0.04\linewidth}
  \begin{subfigure}{0.20\linewidth}
    \centering
    \includegraphics[width=0.75\linewidth]{pictures/Experiment/silk_chiffon_T-shirt.png}
    \subcaption{Chiffon}
  \end{subfigure}
  \caption{Visualization of the three garment-material presets used in the material-robustness evaluation.}
  \label{fig:tshirt-row}
\end{figure}

Evaluation resolutions are aligned with each detector’s standard input size: YOLOv8~\cite{yolov8} and YOLOX~\cite{yolox2021} use \(640\times 640\), SSD300~\cite{SSD} uses \(300\times 300\), and Deformable DETR~\cite{DDETR} uses \(1333\times 800\). For the confidence trajectory experiment, the test sequence is extended to \(327\) frames while keeping all other settings unchanged. Three material presets are instantiated to represent the geometric \(10\%\), \(50\%\), and \(90\%\) points of the training-time material distribution, corresponding to silk/chiffon, cotton, and denim/canvas (Fig.~\ref{fig:tshirt-row}). When a material preset is selected, its physical parameters remain fixed for the entire sequence; all other rendering and simulation settings follow the general protocol. This produces an additional \(432 \times 3 = 1{,}296\) videos.

\noindent\textbf{Physical Test Set:}
Physical garments bearing the adversarial textures were produced using dye-sublimation transfer. Because camera elevation cannot be precisely fixed in real-world environments, the elevation was set to $65^{\circ} \pm 10^{\circ}$. The camera height ranged from 2.5 to $3\,\mathrm{m}$, and the azimuth was sampled from $0^{\circ}$ to $350^{\circ}$ in $30^{\circ}$ increments. At each azimuth, two indoor videos and two outdoor videos were recorded. The initial camera–subject distance was $3\,\mathrm{m}$. Each video contains approximately 180 frames (about $3\,\mathrm{s}$ at $60\,\mathrm{fps}$). Following this protocol, 48 videos were collected for each ablation setting, resulting in a total of 192 videos. All videos were captured at a native resolution of $1920 \times 1080$ (MP4 format). For transferability evaluation, each video was resized to match the native input resolution of each detector before inference, and all results were summarized accordingly.

{
    \small    \bibliographystyle{ieeenat_fullname}
    \bibliography{main}
}

\end{document}